# Opportunities to Parallelize Path Planning Algorithms for Autonomous Underwater Vehicles

Mike Eichhorn
Inst. for Automation and Systems Engineering
Ilmenau University of Technology
98684 Ilmenau, Germany
Email: mike.eichhorn@tu-ilmenau.de

Ulrich Kremer
Dept. of Computer Science
Rutgers University
Piscataway, New Jersey 08854
Email: uli@cs.rutgers.edu

*Abstract* — This paper discusses opportunities to parallelize graph based path planning algorithms in a time varying environment. Parallel architectures have become commonplace, requiring algorithm to be parallelized for efficient execution. An additional focal point of this paper is the inclusion of inaccuracies in path planning as a result of forecast error variance, accuracy of calculation in the cost functions and a different observed vehicle speed in the real mission than planned. In this context, robust path planning algorithms will be described. These algorithms are equally applicable to land based, aerial, or underwater mobile autonomous systems.

The results presented here provide the basis for a future research project in which the parallelized algorithms will be evaluated on multi and many core systems such as the dual core ARM Panda board and the 48 core Single-chip Cloud Computer (SCC). Modern multi and many core processors support a wide range of performance vs. energy tradeoffs that can be exploited in energy-constrained environments such as battery operated autonomous underwater vehicles. For this evaluation, the boards will be deployed within the Slocum glider, a commercially available, buoyancy driven autonomous underwater vehicle (AUV).

## I. INTRODUCTION

The parallelization of programs has gained increasing importance since 2005, when the series production of multi core processors started. Before then, writing parallel software was mainly limited to performance sensitive applications running on high performance servers, distributed computer networks, or supercomputers. To use the full performance of multi and many core processors, programs have to exhibit a significant level of parallelism. This typically requires algorithms to be rewritten or designed with parallelism in mind, which is very different from writing traditional, sequential code. Although harder to program than sequential architectures, multi and many core chips can deliver high performance at low power and energy costs, making the on-board execution of compute and data intensive tasks feasible in battery operated environments such as land based, aerial, or underwater autonomous systems.

Mission planning is an important application for autonomous underwater vehicles (AUVs), where a new route must be calculated during a mission with as little delay as possible using new information in the region of interest. A multi/many core processor, used in an AUV, can do the work of a whole control center on board, which leads to a significant improvement in the fields of application and the effectiveness of AUVs. Moreover, the parallelization of programs also allows new possibilities for the off-line mission planning, where the generated routes can be provided earlier, or an accurate computer intensive glider model [1] as well as additional energy cost models [2] can be included in the planning.

The discussion of opportunities to parallelize graph based path planning algorithms is the main focus of this paper. The sequential versions of these algorithms are described in detail elsewhere [3, 4, 5]. The algorithmical structure of graph methods exposes many opportunities for easy parallelization, including multiple calls to cost functions that calculate the time to travel along an edge, and multiple calls of the entire search algorithm to detect the optimal departure time. In this context, an additional possibility for parallelization will be presented in this paper, namely the inclusion of uncertain information in the path planning. This is an important issue for practice-oriented applications. Every path planning algorithm employs some modeling and/or simulation to verify the quality of a suggested route. These models/simulators cannot "imitate" the real world error-free due to the limited accuracy of the model or simulator and the uncertainty of real vehicle behavior and the region of interest in the future.

There are a variety of approaches to include uncertain information in path planning, especially for mobile autonomous systems. A 3D spatiotemporal grid structure is proposed in [6] to find a mission plan for a Slocum glider [7, 8] in strong, dynamic and uncertain ocean currents. The algorithm uses a wavefront expansion. The inclusion of robot parameter uncertainty in path planning has been discussed in [9]. The algorithm is based on the stochastic response surface method (SRSM). The uncertain robot parameters are the front and rear axle roll stiffness values. A path planning algorithm for unmanned aerial vehicles in uncertain and adversarial environments is described in [10]. The algorithm is based on a probability map of threats used the Bellman-Ford search-algorithm.

This paper presents path planning algorithms for time-varying environments using uncertain information. The algorithms build upon the graph algorithms described in [3]. The algorithms use the principal of a parameter space approach to include uncertain information in the path search. To do this, each examined edge will be simulated multiple times with possible parameter combinations of the uncertain parameters, which are the ocean current components u and v, and the vehicle speed vveh_bf for several start times. The result is a time period within the cost value can lie. This approach offers a considerable potential for parallelization.

The results presented in this paper provide the basis for a future research project which will evaluate the parallelized algorithms on different multi/many core architectures such as the dual core ARM Panda board [11], and the 48 core Single-chip Cloud Computer (SCC), a research microprocessor developed by Intel [12]. In addition to providing raw performance, experimental platforms such as the SCC can support a range of performance vs. energy tradeoffs by allowing their cores, on-chip network, and memory controllers to be dynamically voltage and frequency scaled. In CMOS technology, the dissipated power is proportional to the square of the supply voltage, and frequency and supply voltage are basically directly proportional. As a result, a parallel algorithm that is executed on more, but slower cores can have the same performance, but may consume significantly less energy when executing on the slower cores. This physical property has been exploited before in the context of data centers and supercomputers [13, 14]. However, the experimental SCC system is unique in the sense that it allows to exploit substantial parallelism for energy reduction in low power environments such as the battery operated Slocum glider. We are planning to evaluate the Panda board as well in order to have a power/performance comparison between the SCC and a current, commercially available, low power multi core system.

This evaluation phase will be followed by actual deployments of the Slocum glider with the SCC board off the coast of New Jersey. The path planning algorithms will calculate online an optimal route as promptly as possible, using new information like sea currents, weather conditions, and remaining battery energy. Being able to perform path planning tasks on-board can be more energy efficient since expensive communication with a remote control center is avoided, or may be desirable for missions where such communication may only be intermittent or not possible at all.

## II. OPPORTUNITIES FOR PARALLELIZATION

A parallelization can be realized on multiple levels of the graph based path planning algorithms. Figure 1 shows possible concepts for the parallelization. For example, the detection of the sea current information using netCDF files is parallelizable because the 3–5 sampling points of the interpolation algorithm can be calculated at the same time (see section II.C in [2]). The determination of the optimal dive profile, using 10-20 cost functions to calculate the travel time in the several depth profiles also allows an easy possibility for parallelization (see section III.B in [2]). The inclusion of uncertain information in path planning offers a further opportunity for a multiple call of the cost function. The travel time to drive along an edge will be calculated for all possible combinations of the uncertain parameters: ocean current and vehicle speed. A detailed description of this concept is presented in section III of this paper. The simultaneous processing of the examined neighbour edges in the graph algorithm seems obvious. A synchronistic examination of 5-32 edges here is possible (see section I.E in [3]). The multiple call of the search algorithm to create approximated runs of the curve to detect the optimal departure time (see section III.B in [3]) provides a good opportunity for parallelization.

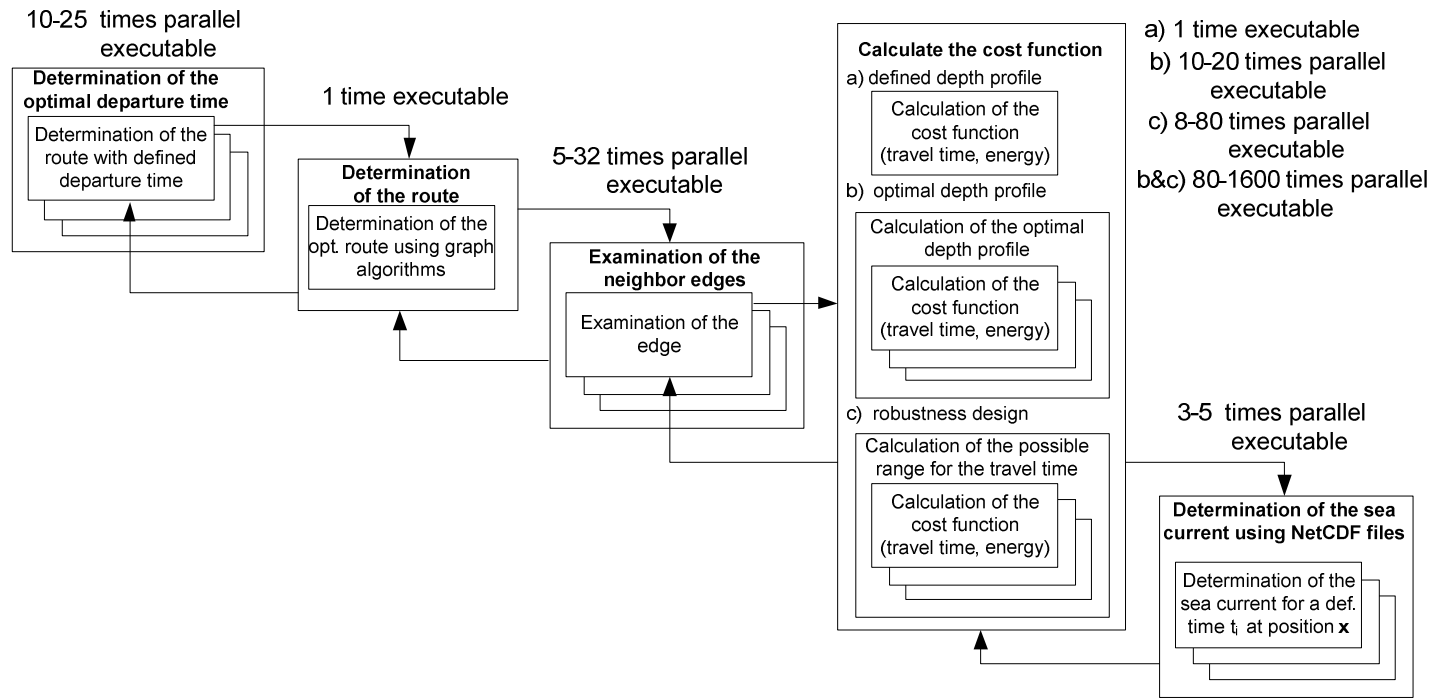

Figure 1 Opportunities to parallelize graph based path planning algorithms

## III. ROBUST DESIGN IN PATH PLANNING

### A. Idea

The idea to design a robust path planning algorithm is based on the parameter space approach in robust control [15]. A representative example in this research field is the position controller design for a crane or a loading bridge where the moved load and the rope length can vary greatly. The load varies between the weight of the empty transport hook and the maximum permissible load. The length of the rope changes when picking up, transporting and setting down the load. The task is to find control parameters for all possible combinations of the uncertain plant parameters mass and rope length. The uncertain parameters $q_i$ will be bounded by an operating domain $Q$, typically a hyper rectangle.

$$Q = \{q \mid q_i \in [q_i^-; q_i^+], i = 1, 2, ..., l\} \quad (1)$$

A pragmatic design approach finds a robust controller for all vertices of the hyper rectangle. In the crane example mentioned above it would be four vertices. If necessary, additional parameter combinations in the inside or on the edges/surfaces of the hyper rectangle $Q$ must be defined. A basic rule in robust control is: "Be an optimist in design and a pessimist in analysis."[15]. This means that each approach is permitted, if afterwards the result will be analysed [16].

The uncertainty in path planning is a result of

- forecast error variance,
- accuracy of calculation in the cost functions and
- a different vehicle speed in the real mission than planned.

This results in three uncertain parameters: the ocean current components $u$ and $v$ and the vehicle speed through water $v_{veh\_bf}$. An easy way to consider all factors responsible for an incorrect cost function calculation in the robustness design is the definition of an error variance for the vehicle speed. The forecast delivers a field of ocean current vectors at defined positions $\mathbf{x_i}$, and times $t_i$ with an error variance of $\pm u$ and $\pm v$. Figure 2 shows the operating domain for a single ocean current $v_c$ with its components $u$ and $v$ that is the result.

The operating domain $Q$ in Figure 3 defines the area wherein the real ocean current vector can lie. There exist a million of possibilities for the size and direction of the current vector $\mathbf{v}_c$. In case of a pragmatic design approach in path planning only four representatives of the ocean current vector $\mathbf{v}_c$ will be used. These are the four vertices of the operating domain $Q$:

$$q_v = \{v_c^{--}, v_c^{-+}, v_c^{+-}, v_c^{++}\} \quad (2)$$

with

$$\begin{aligned} v_c^{--} &= [u - \Delta u \quad v - \Delta v] \\ v_c^{-+} &= [u - \Delta u \quad v + \Delta v] \\ v_c^{+-} &= [u + \Delta u \quad v - \Delta v] \\ v_c^{++} &= [u + \Delta u \quad v + \Delta v] \end{aligned} \quad (3)$$

Figure 3 shows a path consisting of three edges from a start point $s_0$ to a goal point $s_3$ in an uncertain ocean current field. Depending on the real ocean current conditions, the vehicle can reach the goal point $s_3$ sooner or later.

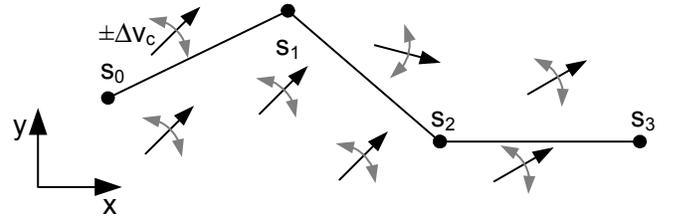

Figure 3: Path through an ocean current field with uncertain information

The approach explained above will be used in a simulation to find the possible arrival times in goal point $s_3$. This is illustrated in Figure 4. The calculation begins at start point $s_0$ at start time $t_0$. Four simulations using the parameter sets of equation (3) will be executed to determine the possible arrival times in way point $s_1$. The result is a time period with a lower $d_1^-$ and an upper $d_1^+$ value. From these two values the next simulations will be started, each with the four parameter sets. This will be repeated for waypoint $s_2$. As a result a time period $[d_3^- \, d_3^+]$ to define the possible arrival time in way point $s_3$ will be created.

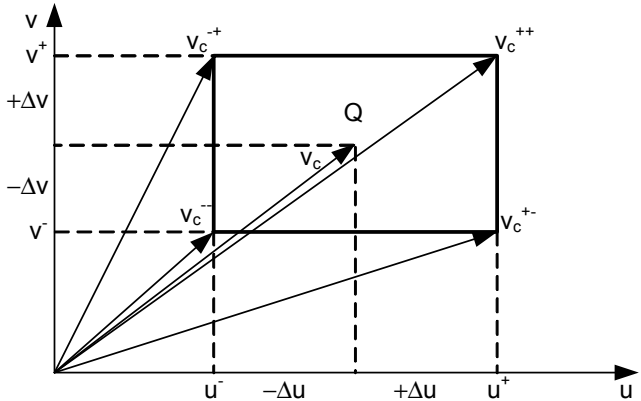

Figure 2: Operating domain $Q$ for the ocean current $v_c$

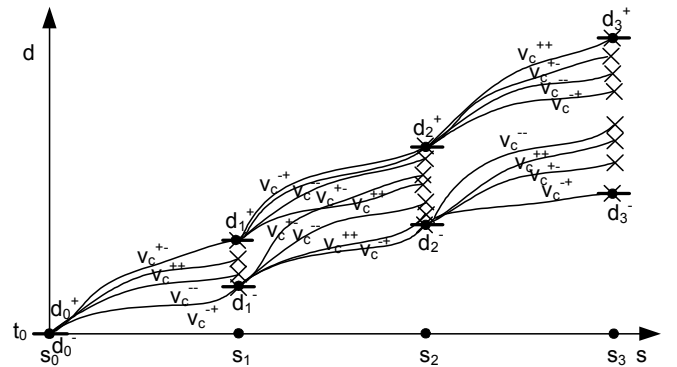

Figure 4: Travel time courses of the several way points $s_i$

## B. Robust Path Planning Algorithms

This section describes the idea of robust path planning algorithms in a time variant environment. The algorithms build upon the TVE (time-varying environment) algorithm and its modifications (A*TVE, ZTVE, ZA*TVE) to accelerate the search. The algorithms are described in detail in [5].

The TVE algorithm will be employed to demonstrate the functional principle on how uncertain information can be included in a robust path planning algorithm. Table I shows a comparison between the TVE algorithm (left column) and the robust TVE (RTVE) algorithm (right column). The shaded text fields highlight the differences between the algorithms. There are the following differences:

1. The robust algorithm (RTVE) uses a lower $d^-$ and upper $d^+$ cost value to define the time period wherein the cost value can lie for each vertex. These values are the result of the uncertain information (see Figure 4 and section D).
2. The selection criterion to examine and to relax the edge $(u, v)$ is the size comparison between the upper cost values $(d^+[u], d^+[v])$ of vertex $u$ and $v$.
3. The RTVE algorithm uses a function $ufunc$ to calculate a time period $[d^-[v]\ d^+[v]]$ wherein the vehicle can arrive at vertex $v$. This leads to parting from vertex $u$ at the departure time period $[d^-[u], d^+[u]]$. The costs $d^-[u]$ and $d^+[u]$ describe the minimum and maximum travel time from the source vertex $s$ to the vertex $u$. The function $ufunc$ will be presented in section D in more detail.
4. The RTVE algorithm uses the upper cost value $d^+$ of the visited vertex $v$ in the priority queue $Q$.

## C. Details of the robust ZA*TVE algorithm

The use of the modified TVE algorithms [5] for a path search with uncertain information is also possible and will be described for the ZA*TVE algorithm subsequently. The other algorithms can be built by omitting the corresponding operators. The necessary modifications can be seen in Table II. The yellow (light grey) shaded text fields highlight the differences of the TVE algorithm (see Table I), the gray shaded fields are the additional modifications between the ZA*TVE algorithm (left column) and the robust TVE (RZA*TVE) algorithm (right column). These modifications are:

1. The function CALC-UNOPTDIR calculates a vector of optimal path directions $\phi_{opt\_vec}$ for all possible combinations of uncertain parameters (see also section II.C in [3]).
2. A successor edge will be selected if its direction corresponds with any optimal path direction $\phi_{opt\_vec}[i]$ within an angle variation $\pm\Delta\phi_{max}$.
3. The function $f[v]$ uses the upper cost value $d^+_v$.

TABLE I PSEUDO-CODE OF THE TVE AND RTVE ALGORITHMS

| TVE($G, s, t_0$) | RTVE($G, s, t_0$) |
|---|---|
| for each vertex $u \in V$ | for each vertex $u \in V$ |
|   $d[u] \leftarrow \infty$ |   $d^-[u] \leftarrow \infty$ |
|  |   $d^+[u] \leftarrow \infty$ |
|   $\pi[u] \leftarrow \infty$ |   $\pi[u] \leftarrow \infty$ |
|   $color[u] \leftarrow$ WHITE |   $color[u] \leftarrow$ WHITE |
| $color[s] \leftarrow$ GRAY | $color[s] \leftarrow$ GRAY |
| $d[s] \leftarrow t_0$ | $d^-[s] \leftarrow t_0$ |
|  | $d^+[s] \leftarrow t_0$ |
| INSERT($Q, s$) | INSERT($Q, s$) |
| while ($Q \neq \emptyset$) | while ($Q \neq \emptyset$) |
|   $u \leftarrow$ EXTRACT-MIN($Q$) |   $u \leftarrow$ EXTRACT-MIN($Q$) |
|   $color[u] \leftarrow$ BLACK |   $color[u] \leftarrow$ BLACK |
|   for each $v \in Adj[u]$ |   for each $v \in Adj[u]$ |
|     if ($d[u] < d[v]$) |     if ($d^+[u] < d^+[v]$) |
|       $d_v = wfunc(u, v, d[u]) + d[u]$ |       $d^-_v, d^+_v = ufunc(u, v, d^-[u], d^+[u])$ |
|       if ($d_v < d[v]$) |       if ($d^+_v < d^+[v]$) |
|         $d[v] \leftarrow d_v$ |         $d^-[v] \leftarrow d^-_v$ |
|  |         $d^+[v] \leftarrow d^+_v$ |
|         $\pi[v] \leftarrow u$ |         $\pi[v] \leftarrow u$ |
|         if ($color[v]$ = GRAY) |         if ($color[v]$ = GRAY) |
|           DECREASE-KEY($Q, v, d_v$) |           DECREASE-KEY($Q, v, d^+_v$) |
|         else |         else |
|           $color[v] \leftarrow$ GRAY |           $color[v] \leftarrow$ GRAY |
|           INSERT($Q, v$) |           INSERT($Q, v$) |
| return ($d, \pi$) | return ($d, \pi$) |

TABLE II PSEUDO-CODE OF THE ZA*TVE AND RZA*TVE ALGORITHMS

| ZA*TVE($G, s, g, t_0, \Delta\phi_{max}$) | RZA*TVE($G, s, g, t_0, \Delta\phi_{max}$) |
|---|---|
| for each vertex $u \in V$ | for each vertex $u \in V$ |
|   $d[u] \leftarrow \infty$ |   $d^-[u] \leftarrow \infty$ |
|  |   $d^+[u] \leftarrow \infty$ |
|   $f[u] \leftarrow \infty$ |   $f[u] \leftarrow \infty$ |
|   $\pi[u] \leftarrow \infty$ |   $\pi[u] \leftarrow \infty$ |
|   $color[u] \leftarrow$ WHITE |   $color[u] \leftarrow$ WHITE |
| $color[s] \leftarrow$ GRAY | $color[s] \leftarrow$ GRAY |
| $d[s] \leftarrow t_0$ | $d^-[s] \leftarrow t_0$ |
|  | $d^+[s] \leftarrow t_0$ |
| $f[s] \leftarrow t_0 + h(s)$ | $f[s] \leftarrow t_0 + h(s)$ |
| INSERT($Q, s$) | INSERT($Q, s$) |
| while ($Q \neq \emptyset$) | while ($Q \neq \emptyset$) |
|   $u \leftarrow$ EXTRACT-MIN($Q$) |   $u \leftarrow$ EXTRACT-MIN($Q$) |
|   if ($u = g$) |   if ($u = g$) |
|     return ($d, \pi$) |     return ($d, \pi$) |
|   $color[u] \leftarrow$ BLACK |   $color[u] \leftarrow$ BLACK |
|   if ($u \neq s$) |   if ($u \neq s$) |
|     $\phi_{opt}$=CALC-OPTDIR($\pi[u], u,$ |     $\phi_{opt\_vec}$=CALC-UNOPTDIR($\pi[u], u,$ |
|       $d[\pi[u]], d[u]$) |       $d^-[\pi[u]], d^+[\pi[u]], d^-[u], d^+[u]$) |
|   for each $v \in Adj[u]$ |   for each $v \in Adj[u]$ |
|     if ($d[u] < d[v]$) |     if ($d^+[u] < d^+[v]$) |
|       $\phi_{path}$ = PATH-PATHDIR ($u,v$) |       $\phi_{path}$ = CALC-PATHDIR ($u,v$) |
|       if (($u=s$) OR |       if (($u=s$) OR |
|         ($\|\phi_{opt}-\phi_{path}\| < \Delta\phi_{max}$)) |         if any($\|\phi_{opt\_vec}[i]-\phi_{path}\| < \Delta\phi_{max}$)) |
|         $d_v = wfunc(u, v, d[u]) + d[u]$ |         $d^-_v, d^+_v = ufunc(u, v, d^-[u], d^+[u])$ |
|         if ($d_v < d[v]$) |         if ($d^+_v < d^+[v]$) |
|           $d[v] \leftarrow d_v$ |           $d^-[v] \leftarrow d^-_v$ |
|  |           $d^+[v] \leftarrow d^+_v$ |
|           $f[v] \leftarrow d_v + h(v)$ |           $f[v] \leftarrow d^+_v + h(v)$ |
|           $\pi[v] \leftarrow u$ |           $\pi[v] \leftarrow u$ |
|           if ($color[v]$ = GRAY) |           if ($color[v]$ = GRAY) |
|             DECREASE-KEY($Q, v, f[v]$) |             DECREASE-KEY($Q, v, f[v]$) |
|           else |           else |
|             $color[v] \leftarrow$ GRAY |             $color[v] \leftarrow$ GRAY |
|             INSERT($Q, v$) |             INSERT($Q, v$) |
| return ($d, \pi$) | return ($d, \pi$) |

## D. Robust cost function

A graph based path planning algorithm needs a cost value $d$ to examine the several edges during a search. This value corresponds to the travel time from the start vertex $s$ to the examined vertex. By inclusion of uncertain information in the planning this single cost value $d$ can be varied within a range defined by a lower an upper boundary $[d^- \; d^+]$. The calculation of this range is based on the parameter space approach, as described in section A. The uncertain parameters are the ocean current components $u$ and $v$ and the vehicle speed through water $v_{veh\_bf}$ (cruising speed) with its error variances $\pm\Delta u$ $\pm\Delta v$ and $\pm\Delta v_{veh\_bf}$. This leads to eight parameter sets using a pragmatic design approach which describe possible deviations of the notional conditions for $u$, $v$, and $v_{veh\_bf}$ (see Figure 5).

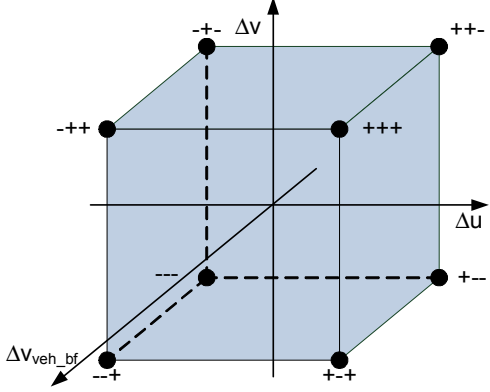

Figure 5: Parameter sets using the vertices of the hyper rectangle

Each of the eight parameter sets will be included in a weight function which calculates the travel time to drive along the edge from a start vertex $u$ to an end vertex $v$ using a given start time $t_{start}$. (This function is described in detail in section III in [1].) For every call of the weight function, another parameter set $q_v[i]$ will be used to simulate its influence on the calculated travel time. The start time $t_{start}$ is the same for all eight calculations. The easiest way to determine the time period of the cost value of vertex $v$ is the multiple calculation of the weight function using the eight parameter sets for the lower $d^-$ and the upper $d^+$ cost value boundary of vertex $u$ as start time $t_{start}$ (see also Figure 4). The time period for the cost value $[d_v^- \; d_v^+]$ of vertex $v$ is the result of the shortest and the longest arrival time in vertex $v$ of all calculations:

$$d_v = \begin{bmatrix} \forall i \; wfunc(u,v,d^-[u],q_v[i]) + d^-[u], \\ \forall i \; wfunc(u,v,d^+[u],q_v[i]) + d^+[u] \end{bmatrix} \quad (4)$$

$$[d_v^- \; d_v^+] = [\min(d_v) \; \max(d_v)]$$

Depending on the time gradient of the time-varying ocean current and the size of the time period additional start times $t_{start\_1} \ldots t_{start\_n}$ within the time period must be defined:

$$d_v = \begin{bmatrix} \forall i \; wfunc(u,v,d^-[u],q_v[i]) + d^-[u], \\ \forall i \; wfunc(u,v,t_{start\_1},q_v[i]) + t_{start\_1}, \ldots \\ \forall i \; wfunc(u,v,t_{start\_n},q_v[i]) + t_{start\_n}, \\ \forall i \; wfunc(u,v,d^+[u],q_v[i]) + d^+[u] \end{bmatrix}. \quad (5)$$

## IV. RESULTS

### A. The selected test function for a Time-Varying Ocean Flow

The function used to represent a time-varying ocean flow describes a meandering jet in the eastward direction, which is a simple mathematical model of the Gulf Stream [17] and [18]. This function was applied in [3, 4 and 5] to test the TVE algorithm and its modifications and will be used in the following sections to show the influence of the methods to realize fast search algorithms and to find suboptimal paths using uncertain information. The stream function is:

$$\phi(x,y) = 1 - \tanh\left( \frac{y - B(t)\cos(k(x-ct))}{\left(1 + k^2 B(t)^2 \sin^2(k(x-ct))\right)^{\frac{1}{2}}} \right) \quad (6)$$

which uses a dimensionless function of a time-dependent oscillation of the meander amplitude

$$B(t) = B_0 + \varepsilon \cos(\omega t + \theta) \quad (7)$$

and the parameter set $B_0 = 1.2$, $\varepsilon = 0.3$, $\omega = 0.4$, $\theta = \pi/2$, $k = 0.84$ and $c = 0.12$ to describe the velocity field:

$$u(x,y,t) = -\frac{\partial \phi}{\partial y} \quad v(x,y,t) = \frac{\partial \phi}{\partial x}. \quad (8)$$

The dimensionless value for the body-fixed vehicle velocity $v_{veh\_bf}$ is 0.5. This test function makes it possible to show very transparently how a path planning algorithm works with uncertain information. The exact time optimal solution without uncertain information was found by solving a boundary value problem (BVP) with a collocation method bcp6c [19] in MATLAB. For more information see [5].

### B. Comparison between the several robust path planning algorithms

This section presents the results using the robust path planning algorithms, which are described in section III.B. For the test cases, five different start positions were distributed in the whole area of operation as shown in Figure 6. An error variance of 5% for the vehicle speed $v_{veh\_bf}$ and the ocean current components $u$ and $v$ will be used in the tests to include the uncertain information in the path planning. Figure 6 shows the five paths found using optimal control and the robust graph methods. All graph methods found the same paths, why only one path per start point will be plotted. The necessary number of robust cost function calls (RCFC), cost function calls (CFC) and current model calls (CMC) are shown in TABLE III. Figure 7 shows the number of cost function calls (CFC) using the robust graph methods for the five start positions. The use of the several methods to accelerate the robust TVE algorithm (RTVE) is directly reflected in a decrease of cost function calls. Using the A* algorithm in a robust graph method, the number of function calls correlates directly with the distance between the start and the goal point. This is reasonable since the algorithm includes only a subset of the vertices in the path search, in fact, only the preferred vertices with a short distance to the goal point. The use of the RZA*TVE algorithm allows a decrease of the number of cost function calls by about a factor of 6 to 200 in comparison to the base TVE algorithm. The reason for this enormous decrease is the pre selection of possible successor edges during the search (see Table II and section II.C in [5]). The RTVA algorithm examined all possible paths. With increasing travel time, the time period for the costs of the several vertices will also increase. This leads to the definition of additional start times within the time period and thus to additional cost function calculations.

## C. Influence of error variance in uncertain parameters

An analysis of the error variance influence of the uncertain parameters to the found paths will be presented in this section. The examined error variance is 0%, 2.5%, 5%, 10% and 15% for the vehicle speed $v_{veh\_bf}$ and the ocean current components $u$ and $v$. For the test cases, start position SP1 and SP2 in Figure 8 will be used. In case of start position SP1, the robust path planning algorithms can find a route for error variances up to 5 % (see Figure 8). In case of larger variances, there do not exist routes to cross the strong sea current stream in the mainstream of jet. For start position SP2, the robust algorithms found a path for every variance. The algorithms used less and less the mainstream of jet with rising variances. At the beginning of the paths, where the time period of the cost functions is smaller (see Figure 9), the paths found are close to the optimal solution. With increasing uncertain information the planning algorithms avoid areas where an adverse current may exist and favor a safe route to the goal point.

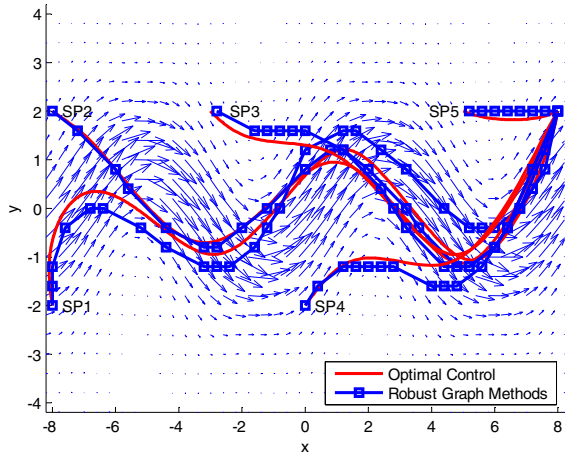

Figure 6: Determined paths though a time-varying ocean field using Optimal Control and the Robust Graph Methods for different start positions and uncertain information of 5% error variance

TABLE III
RESULTS OF THE DIFFERENT SEARCH METHODS

| Method | SP1 No. of RCFC No. of CFC No. of CMC | SP2 No. of RCFC No. of CFC No. of CMC | SP3 No. of RCFC No. of CFC No. of CMC | SP4 No. of RCFC No. of CFC No. of CMC | SP5 No. of RCFC No. of CFC No. of CMC |
|---|---|---|---|---|---|
| RTVE | 11924 443816 3102942 | 12007 316272 2148223 | 11589 461488 3339934 | 11970 507232 3670745 | 11508 675568 5123954 |
| RA*TVE | 7610 223608 1451654 | 6954 150944 887092 | 4275 103872 626605 | 3176 75920 425213 | 688 11024 93602 |
| RZTVE | 3281 104056 905125 | 3639 90784 791006 | 2356 65312 576344 | 1663 45424 395350 | 388 7056 63263 |
| RZA*TVE | 2642 77648 696935 | 2362 51800 459393 | 1381 33392 291697 | 991 23280 191192 | 200 2976 28392 |

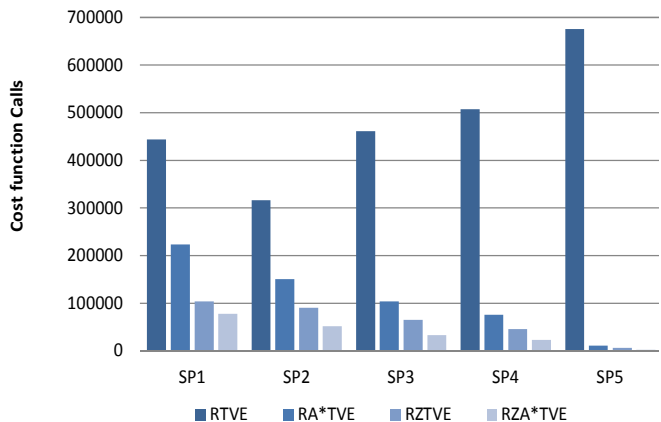

Figure 7 Cost function calls for the various robust graph methods with different start positions

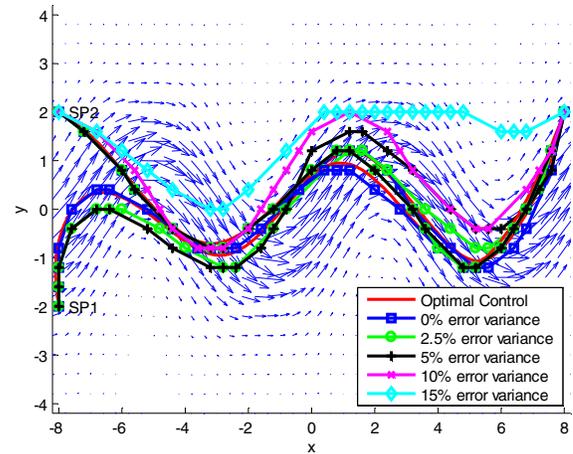

Figure 8: Determined paths though a time-varying ocean field using different error variances in the uncertain information

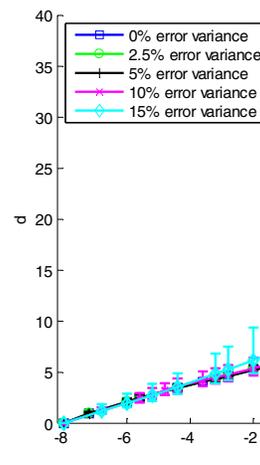

Figure 9: Time periods of the cost function $d$ for the serveral way points of the found paths for SP2 using different error variances in the uncertain information

## V. Conclusions and Future Work

In this paper, different opportunities are presented to parallelize graph based path planning algorithms in a time varying environment. The described path planning algorithms for time- varying environment using uncertain information was tested with an analytical time-variant function. The results are plausible and comprehensible.

The first benchmark tests of the algorithms, which are written in Microsoft Visual Studio 2010 by using the Task library [20], which is a component of the Boost Sandbox [21] are currently underway. An Intel® Xeon® Processor E7 with 10 cores is the heart of this first test platform. Once the code is stable, we will port it onto the Panda and SCC systems available in the EEL research lab at Rutgers, followed by an extensive study of possible performance vs. energy tradeoffs using physical power measurements. As the final step, we will use one of the two Slocum gliders in the EEL lab as our physical deployment test bed, with several test missions planned off the coast of New Jersey.

## VI. Acknowledgement

We would like to thank Intel for providing the EEL lab at Rutgers with an SCC system in support of this research. We are particularly grateful to the SCC development team for their help and advice.